\documentclass[10pt]{article}
\usepackage{amsmath}
\usepackage{amsfonts}
\usepackage{amssymb}
\usepackage{amsthm}

\usepackage[onehalfspacing]{setspace}
\usepackage[margin = 0.75 in]{geometry}
\usepackage{graphicx}
\graphicspath{{./figures/}}
\usepackage{indentfirst}
\usepackage{algorithm}
\usepackage{algorithmic}

\usepackage{mathtools}

\usepackage{booktabs}

\usepackage{subcaption}

\usepackage[section]{placeins}

\usepackage{bbm}

\usepackage{authblk}

\usepackage{hyperref}
\hypersetup{
    colorlinks,
    citecolor = blue,
    filecolor = blue,
    linkcolor = blue,
    urlcolor = blue,
}

\usepackage{xcolor}
\newcommand{\tableformat}{\fontsize{9}{10}\selectfont
\centering}

\definecolor{xblue}{RGB}{46, 77, 167}

\newcommand{\best}{\boldsymbol}
\newcommand{\second}{\color{xblue}\boldsymbol}
\newcommand{\select}{\boldsymbol}

\usepackage{natbib}

\newcommand{\imagewidth}{0.45\textwidth}
\newcommand{\subwidth}{\textwidth}

\newcommand{\code}[1]{\texttt{#1}}

\newcommand{\name}{MARVEL}
\newcommand{\CE}{CE}

\DeclareMathOperator*{\argmax}{argmax}
\DeclarePairedDelimiterX{\brac}[1]{[}{]}{#1}
\newcommand{\indic}[1]{\mathbbm{1}\brac*{#1}}
\newcommand{\real}{\mathbb{R}}
\newcommand{\sign}{\text{sign}}

\newcommand{\iton}{_{i=1}^n}
\newcommand{\jton}{_{j=1}^n}
\newcommand{\jtok}{_{j=1}^k}
\newcommand{\jtom}{_{j=1}^m}

\renewcommand{\leq}{\leqslant}
\renewcommand{\geq}{\geqslant}
\renewcommand{\subset}{\subseteq}
\newcommand{\gap}{\quad}
\renewcommand{\hat}{\widehat}

\newcommand{\iiib}{}

\begin{document}

\title{Learning to Combat Noisy Labels via Classification Margins}
\author[1]{Jason Z. Lin\thanks{jasonlin@ucsd.edu}}
\author[1,2]{Jelena Bradic\thanks{jbradic@ucsd.edu}}
\affil[1]{Department of Mathematics, University of California, San Diego}
\affil[2]{Halicio\v{g}lu Data Science Institute, University of California, San Diego}
\date{}
\setcounter{Maxaffil}{0}
\renewcommand\Affilfont{\itshape\small}
\maketitle

\begin{abstract}

A deep neural network trained on noisy labels is known to quickly lose its power to discriminate clean instances from noisy ones. After the early learning phase has ended, the network memorizes the noisy instances, which leads to a significant degradation in its generalization performance. To resolve this issue, we propose \name{} (MARgins Via Early Learning), a new robust learning method where the memorization of the noisy instances is curbed.
We propose a new test statistic that tracks the goodness of ``fit''  of every instance {\iiib based on} the epoch-history of its classification margins. If {\iiib its classification margin} is small in a sequence of consecutive learning epochs, that instance is declared noisy and the network abandons learning on it. Consequently, the network first flags a possibly noisy instance, and then waits to see if learning on that instance can be improved and if not, the network learns with confidence that this instance can be safely abandoned. We also propose \name+, where arduous instances can be upweighted, enabling the network to focus and improve its learning on them and consequently its generalization.
Experimental results on benchmark datasets with synthetic label noise and real-world datasets show that \name{} outperforms other baselines consistently across different noise levels, with a significantly larger margin under asymmetric noise.
\end{abstract}

\section{Introduction}

In recent years, deep neural networks have emerged as the state-of-the-art method for classification in numerous domains, in particular computer vision \citep{krizhevsky2012, zeiler2014, simonyan2015, szegedy2015, he2016}. However, a large-scale dataset with clean annotations, such as ImageNet \citep{deng2009}, is usually required for the networks to show superior performance. In practice, obtaining such a large amount of data with manually verified labels is prohibitively laborious or expensive. On the other hand, various crowdsourcing platforms make it possible (and cheaper) to collect large amounts of annotated data, even though the labels may be incorrect. This calls for robust methods that can still effectively learn from the copious data in the presence of noisy labels.

Despite the difficulty in generalization, in the presence of noisy labels, deep neural networks do exhibit the so-called early learning phenomenon \citep{liu2020}, in which the clean instances are fit to the network before the noisy ones are (over-)fit to the network; see e.g., \citet{zhang2017, arpit2017}. On the theoretical front, \citet{liu2020} show that even simple linear models in high dimensions exhibit such behavior as well.  
 This intriguing phenomenon, as a double-edged sword, begs the question: Is there a way to leverage the network's discriminating power, obtained during the early stage of the training, to prevent memorization of the noisy labels at the later stage of the training?

In this paper, we propose \name{} (MARgins Via Early Learning), a method that attempts to robustly identify and exclude noisy instances from future training by designing a new epoch-history driven test statistic that tracks the distributional behavior of the classification margins. This idea of removing noisy instances resonates with the observation in \citet{lapedriza2013} that not all training instances are equally valuable, especially true in the context of learning from noisy labels. Some are difficult but extremely useful for nuanced learning, of crucial importance for strong generalization. If such instances are discarded prematurely, the network will lose some of its generalization property. Instead, \name{} allows focused learning on such instances. In \name+ we introduce an adaptive weighting scheme that automatically adjusts the instance weights based on the historical classification margins. The network assigns sequentially higher weights to difficult instances over the training epochs, whereas noisy instances are discarded or downweighted epoch by epoch.


\name{} is closely connected to and adds to the recent but growing literature on algorithms that identify and filter out suspected noisy instances. Iterative Noisy Cross-Validation (INCV) \citep{chen2019} trains two networks via cross-fitting, and removes a fraction of the highest-loss instances during prediction. This method resembles Co-teaching \citep{han2018} (see below), and may share the same limitations.
DY-Bootstrap \citep{arazo2019} models the loss distributions of the clean and the noisy instances using a beta mixture model, and categorizes each instance according to the posterior probability of being clean or noisy.
Deep Abstention Classifier (DAC) \citep{thulasidasan2019} adds one more class, the class of abstention, to the original $k$-class classification problem, and adds a regularization term that penalizes abstention to the normalized cross entropy loss. However, DAC requires a clean validation set. \citet{song2020-1} calculate a threshold based on the losses of (simulated) instances under symmetric noise, and then remove instances with a loss higher than this threshold. However, such a threshold may be less effective under asymmetric noise, where the losses of clean and noisy instances are less distinct. Finally, Area Under the Margin (AUM) Ranking \citep{pleiss2020} calculates the average of the margins up to the first learning rate decay, and then identifies mislabeled instances with average margins under a particular data-dependent threshold.

While both AUM and \name{} make use of margins to detect noisy instances, they differ in several important aspects. First, AUM is a less flexible algorithm. AUM builds on the observation that the AUM statistic is different for clean and noisy instances under symmetric noise. \name{} adapts to the noise distribution as it learns whether or not an instance is noisy for a few sequential epochs, and then only gives up if its test statistic is {\iiib consistently small}. Moreover, \name's test statistic is more robust to the asymmetric noise distributions.
AUM identifies a noisy instance by comparing its average margin with a given threshold, whereas \name{} uses the maximum margin statistic and compares it  with zero. Lastly, \name{} removes instances at every epoch after the warm-up period, whereas other methods described above (including AUM) perform noise filtering once and for all. In other words, \name{} naturally corresponds to an on-the-fly, progressive noise cleansing approach, in which learning and noise filtering go hand in hand. We believe such an alternating scheme is more beneficial in terms of prediction accuracy as well as noisy label detection.

{\iiib In addition}, there is an evolving line of work on the early learning phenomenon and early stopping methods.
\citet{ma2018} observe the dimensionality compression and expansion stages of training a deep neural network, and propose an adaptive loss function according to the subspace dimensionality during training.
\citet{han2018} propose Co-teaching, where two networks select for each other the training instances to learn from. However, the selected instances typically have low losses, and these easy instances may not help the network generalize to unseen data.
\citet{song2020} propose to train a network with early stopping, and then resume the training using only instances that are correctly classified. Their method suffers from the same drawbacks as \citet{han2018}.
Theoretically, \citet{li2020} show that when training an over-parameterized network, first-order early stopping methods are provably robust to noisy labels. However, in practice, we observe that for a network trained with cross entropy loss, its best test accuracy is still inferior to that of other robust methods (for example, see Figure~\ref{01}). Simply using early stopping does not yield optimal performance.
\citet{liu2020} prove that early learning and memorization occur even in high-dimensional linear models, and design a regularization term that magnifies the gradients of clean instances and diminishes the gradients of noisy instances.


\name+ continues learning after the early stopping, by magnifying the importance of difficult instances and reducing the importance of easy instances. This continuation is advantageous in improving the network generalization.

\subsection{Related Work}



Numerous robust loss functions have been proposed to combat label noise. They achieve this goal by mitigating the impact of noisy instances on the loss or on the gradient updates.
In the binary case, \citet{natarajan2013} propose ways to modify any surrogate loss function to make it robust to label noise. However, their method is not specifically designed for deep neural networks and may engender difficulty in practice.
\citet{ghosh2017, ghosh2015} derive sufficient conditions for a robust loss function in the multi-class setting, where training using the resulting loss function on noisy data yields the optimal classifier trained on clean data. Several theoretically founded robust loss functions are then proposed. In the same paper, \citet{ghosh2017} propose mean absolute error (MAE) as a robust loss function for deep learning. However, training with MAE suffers from slower convergence and accuracy drops, especially with challenging datasets like ImageNet \citep{zhang2018}.
Bi-tempered logistic loss \citep{amid2019} introduces two temperature hyperparameters into the logarithm and exponential of cross entropy, effectively making it bounded and tail-heavy.
\citet{shu2020} propose a framework that automatically tunes the robust loss functions by alternately updating the hyperparameters and parameters of the network.
Generalized cross entropy (GCE) \citep{zhang2018} achieves a desirable compromise between MAE and cross entropy. In particular, GCE combines the robustness of MAE, and the fast convergence and accuracy of cross entropy.
Symmetric cross entropy \citep{wang2019} tries to tackle the bottleneck problem that the network still underlearns hard classes while easy classes already start to overfit. Their method includes a regularization term that swaps the ground-truth probability vector and the predicted probability vector in the cross entropy term.

Another approach is sample reweighting methods.
In the binary case, \citet{liu2015} prove a weighting scheme that makes any surrogate loss function robust to label noise. However, their method requires accurate estimates of the noise flipping probabilities, which is in generally hard to obtain.
\citet{ren2018} propose a meta-learning paradigm, where the sample weights at each epoch are derived from gradient descents on a balanced and noise-free validation set. However, such a validation set is sometimes unavailable.
\citet{mirzasoleiman2020} propose an algorithm CRUST, which selects weighted subsets of the training set with a low-rank Jacobian spectrum, whose low dimensionality helps the network learn fast and generalize well.
\citet{huang2020} propose self-adaptive training, which reweights an instance with its maximal class probability, an indication of the network's confidence in each instance. However, their weighting scheme is prone to noisy instances that are classified confidently, upweighting these noisy instances when it shouldn't.


\name{} prevents overfitting on the noisy instances, where the weights can ``keep'' or ``kill'' an instance during training, preventing their undesirable comeback. \name+ considers weights that can ``upweight'' or ``downweight'' the importance of an instance therefore providing a smoother alternative to a ``keep'' or ``kill'' action. \name{} tracks the goodness of fit of each instance, by observing margins over the epoch-history of gradient iterations. If a recent history is suggestive of either a continuous mis-fit or a modest-fit, an instance is down- or up-weighted, respectively.

Sample reweighting is related to curriculum learning \citep{bengio2009}, the idea to learn from easier instances first and gradually including harder ones, thereby forming a curriculum.
Self-paced learning \citep{kumar2010} defines easiness using the loss value of an instance, and controls the learning pace by means of an evolving regularizer.
Later, extensions of self-paced learning are proposed, such as self-paced learning with diversity \citep{jiang2014}, which advocates the use of easy and diverse instances, and self-paced curriculum learning \citep{jiang2015}, which combines prior knowledge as in curriculum learning and the learner's feedback as in self-paced learning.
Numerous other curriculums are proposed as well, such as active bias \citep{chang2017}, and self-paced co-training \citep{ma2017}.
MentorNet \citep{jiang2018} attempts to learn an optimal data-driven curriculum with an auxiliary neural network, and uses it to train a StudentNet. However, to effectively learn a curriculum, their method requires a noise-free validation set at least 10\% of the noisy training set, which is sometimes a luxury.


\name+ operates in the opposite direction, learning to discard less useful instances first and finetune on the useful and difficult instances later. In addition, \name{} does not require the knowledge of noise rates or any noise-free validation set.

\section{Preliminaries}

\subsection{Problem Setup}

In this paper, we first consider the class-conditional random noise model \citep{natarajan2013} in the binary classification case. Let $\mathcal{X}\subset\real^d$ be the feature space, and let $\mathcal{Y}=\{\pm1\}$ be the label space. Let $D=\{(X_i,Y_i)\}\iton$ be an i.i.d. sample drawn from some distribution $\mathcal{D}$ on $\mathcal{X}\times\mathcal{Y}$, where the ground-truth labels $Y_i$'s are assumed unobservable. Let $D_{\rho}=\{(X_i,\tilde{Y_i})\}\iton$ be the noisy training set, where $\tilde{Y_i}$'s denote the (observable) noisy labels. The class-conditional random noise model assumes that the noisy label is independent of the training instance, conditioned on the ground-truth label. That is, for each $i$,
\[P(\tilde{Y}_i=j|X_i=x_i,Y_i=k)=P(\tilde{Y}_i=j|Y_i=k),\gap\forall j,k\in\{\pm1\},\gap\forall x_i\in\mathcal{X}.\]
This model is commonly studied in the literature; otherwise, we need to use inverse propensity weighting \citep{rosenbaum1983}. The random label noise is generated independently for each $i$, such that
\[P(\tilde{Y_i}=+1|Y_i=-1)=\rho_{-1},\gap P(\tilde{Y_i}=-1|Y_i=+1)=\rho_{+1}.\]
The noise levels $\rho_{-1},\rho_{+1}$ are assumed unknown, but satisfy $\rho_{-1}+\rho_{+1}<1$. The noise is called symmetric if $\rho_{-1}=\rho_{+1}$, or otherwise it is asymmetric.

Let $f:\mathcal{X}\to\real$ be a discriminator function. In deep learning, $f$ is typically a deep neural network minus the softmax output layer. The output $f(x)$ is called logit, the predicted class is $\sign(f(x))$, and the predicted probabilities of class $+1$ and $-1$ are, respectively,
\begin{equation}\label{prob-logit}
p_{+1}(x)=\dfrac{e^{f(x)}}{e^{f(x)}+1},\gap p_{-1}(x)=\dfrac{1}{e^{f(x)}+1}.
\end{equation}
The classification margin is defined as $m(x,y)=yf(x)$, with the property that a correctly classified instance has a positive margin $m(x,y)>0$, whereas a misclassified instance has a negative margin $m(x,y)<0$.

A loss function commonly used in training deep neural networks is the cross entropy loss (\CE)
\begin{equation}\label{cross-ent}
L(x,y)=-\dfrac{1}{n}\sum\iton\log p_{y_i}(x_i),
\end{equation}
where $p_{y_i}(x_i)$ is the predicted probability of $y_i$. In the binary case, using~\eqref{prob-logit}, it can be shown that the cross entropy loss~\eqref{cross-ent} reduces to the logistic loss
\begin{equation}\label{logis-loss}
L(x,y)=\dfrac{1}{n}\sum\iton\log\brac*{1+e^{-y_i f(x_i)}}.
\end{equation}

The goal is to find a discriminator $f$ that predicts the ground-truth label of an unseen $X\in\mathcal{X}$, relying only on the noisy training set $D_{\rho}$. The optimal discriminator is the Bayes rule $\gamma=\sign(P(Y|X)-1/2)$, with the unknown $P(Y|X)$.


A popular metric to measure the difficulty of an instance is its loss \citep{kumar2010}: an instance with a higher loss is considered more difficult. With logistic loss~\eqref{logis-loss}, a higher loss amounts to a lower margin. Henceforth we call an instance $x_i$ noisy if its observable label $\tilde{y_i}$ disagrees with its ground-truth label $y_i$; otherwise, this instance $x_i$ is clean. Qualitatively, we call an instance easy if it is \emph{clean} with a large classification margin, and an instance difficult if it is \emph{clean} with a small classification margin (either positive or negative).

\subsection{Memorization Effect of Cross Entropy}



\begin{figure}[ht]
  \begin{subfigure}[t]{\imagewidth}
    \includegraphics[width=0.98\subwidth]{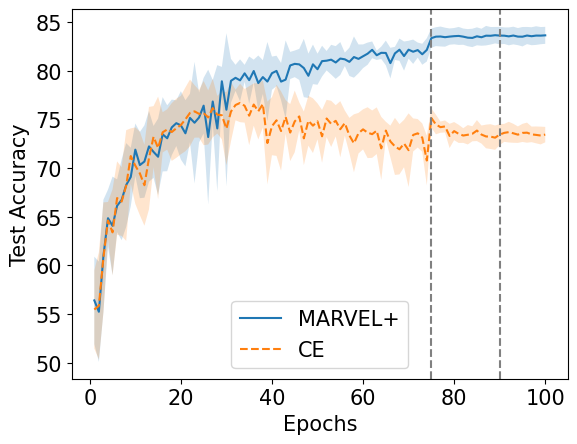}
    \caption{}
    \label{01}
  \end{subfigure}\hfill
  \begin{subfigure}[t]{\imagewidth}
    \includegraphics[width=\subwidth]{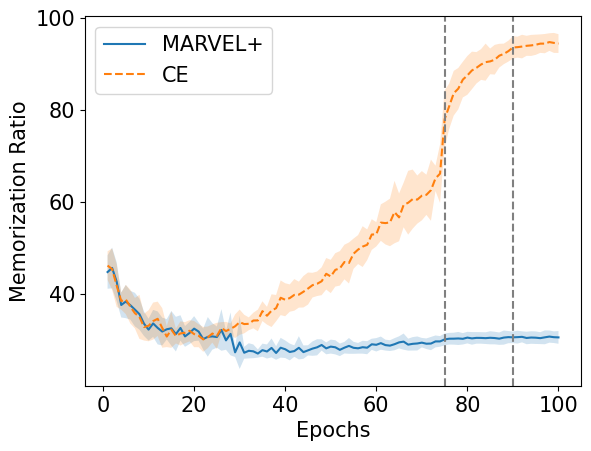}
    \caption{}
    \label{02}
  \end{subfigure}
  \caption{(a) Test accuracy, and (b) memorization ratio~\eqref{mem-ratio} with \name+ (blue) and CE (orange) on CIFAR-10 (cat, dog) with 20\% symmetric noise. The dashed lines are epoch 75 and 90. The mean (curve) and the standard deviation of 10 runs (band) are reported.}
  \label{01-all}
\end{figure}

When training a neural network using \CE, the noisy labels do not prevent the network from learning a coarse representation of the data (they do not help in this respect, either). It is toward the end of the training that the noisy labels begin to bring in significant difficulties for the network and degrade its performance. As empirical evidence, Figure~\ref{01-all} displays the curves of test accuracy and memorization ratio, defined below in~\eqref{mem-ratio},  for \CE. We refer the reader to Section~\ref{subsec-datasets} for a detailed description of the experimental procedures. The experiment is repeated in 10 runs.


From Figure~\ref{01}, we observe that the network trained using \CE{} traverses three phases: the learning phase (epoch 1--20), the apex phase (epoch 20--40), and finally the memorization phase (epoch 40--100). In the learning phase, the network can still pick up signals and learn from the training data, despite the presence of noisy labels.
In the apex phase, the test accuracy stops improving: the network is not able to generalize further, because the gradient signals of the clean instances vanish, whereas the gradient signals of the noisy instances keep increasing \citep{liu2020}. Finally, in the memorization phase, the noisy labels prevail, the network memorizes the noisy labels, driving down its test accuracy. This three-phase process of training a deep neural network using \CE{} is in line with the observation from \citet{zhang2017, arpit2017} that deep neural networks learn clean and easy instances first before memorizing the mislabeled ones. A similar phenomenon is also observed in \citet{ziyin2020}.

Another metric to evaluate the degree of overfitting is the memorization ratio, which we define as the fraction of noisy instances that are fitted to their noisy labels. In equations, if $\hat{Y}$ denotes the classification of an instance $(X,\tilde{Y})$, the memorization ratio is defined as
\begin{equation}\label{mem-ratio}
\dfrac{\#\{(X,\tilde{Y}):\tilde{Y}\neq Y, \hat{Y}=\tilde{Y}\}}{\#\{(X,\tilde{Y}):\tilde{Y}\neq Y\}},\gap (X,\tilde{Y})\in D_{\rho}.
\end{equation}
Note that this metric can be calculated only if the oracle noise information (e.g., which instances are clean, which are noisy) is available. From Figure~\ref{02}, we can also identify the three phases described above. In the learning phase, the network is able to learn from the clean instances while seemingly disregarding the noisy ones, pushing the noisy accuracy down. In the apex phase, we observe that the network already starts to memorize the noisy labels. In the memorization phase, the network continues to memorize the noisy labels, but in an increasingly violent fashion. The learning rate decay at epoch 75 unfortunately enables the network to memorize a lot more noisy labels. A similar phenomenon is observed in \citet{liu2020}.

\section{\name{} and \name+}


We present \name{} in Section~\ref{subsec-remove-noisy} and introduce an extension, \name+, in Section~\ref{subsec-adapt-weight}.

\subsection{Removing Noisy Instances}\label{subsec-remove-noisy}




To remove noisy instances from the training set, it is essential to first distinguish noisy instances from clean ones effectively. We propose to make use of the representation the network learns at the end of the learning phase. This representation is not perfect, but it is useful enough to distinguish noisy and clean instances.

When the network transitions from the learning phase to the apex phase, it has learned an approximately correct representation of the features corresponding to different labels. At this time, a clean training instance is more likely to be correctly classified, whereas a noisy one misclassified. As evidenced by the memorization ratio in Figure~\ref{02}, at the start of the apex phase (epoch 20), the majority of the noisy instances are misclassified.


As the network traverses gradually the apex phase, we expect this discriminating power of the network to last for several epochs. Empirically, from Figure~\ref{02}, we observe that the memorization ratio hits the minimum at the start of the apex phase, and stays roughly at that level for around 10 epochs. This means that the network is still capable of correctly classifying most of the clean instances and misclassifying noisy ones during these 10 epochs.

\begin{figure}[ht]
  \begin{subfigure}[t]{\imagewidth}
    \includegraphics[width=\subwidth]{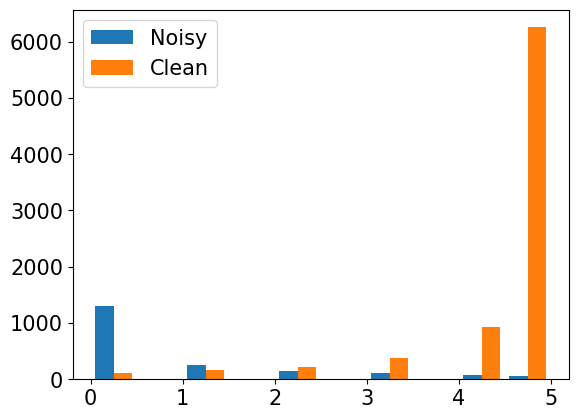}
    \caption{}
    \label{03}
  \end{subfigure}\hfill
  \begin{subfigure}[t]{\imagewidth}
    \includegraphics[width=\subwidth]{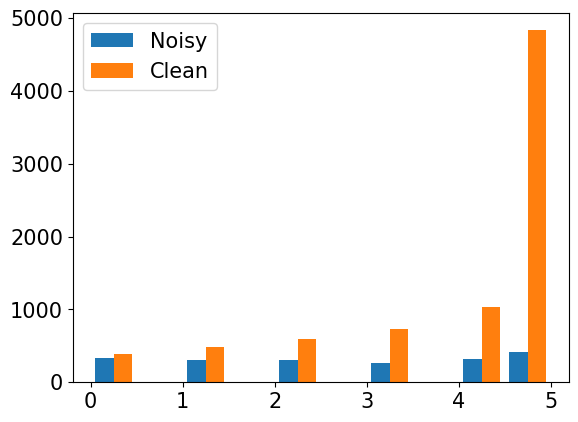}
    \caption{}
    \label{04}
  \end{subfigure}
  \caption{Number of instances with positive margins with noisy (blue) and clean (orange) labels, during the first 5 apex epochs, on CIFAR-10 (a) (truck, car) with 20\% symmetric noise, and (b) (cat, dog) with 40\% asymmetric noise.}
  \label{02-all}
\end{figure}

At the start of the apex phase, the network is expected to consistently misclassify a noisy instance, indicated through a sequence of consecutive negative margins. Figure~\ref{02-all} shows the numbers of positive margins for clean and noisy instances on different datasets. As shown in Figure~\ref{03}, the criterion of consecutive negative classification margins is effective in discerning clean instances from noisy ones: the vast majority of the noisy instances are consistently misclassified, whereas the clean ones are predominantly correctly classified. For the more challenging dataset in Figure~\ref{04}, still roughly 1/6 of the noisy instances are identified, at the expense of a negligible fraction of the clean instances.

Hence, after the network has entered the apex phase, we propose to identify an instance as noisy if it has been misclassified several epochs in a row, and remove it from the training set henceforth. The network is trained via alternately updating the network parameters and the instance weights. Concretely, let $t$ be the epoch, $\theta_t$ be the network parameters at epoch $t$, $w_{j,t}$ be the weight of instance $j$ at epoch $t$, and $\alpha$ be the learning rate. The (full-batch) gradient descent for training the network is
\[\theta_{t+1}=\theta_t-\alpha\dfrac{1}{\sum\jton w_{j,t}}\sum\jton w_{j,t}\nabla L_{\theta_t}(x_j,y_j).\]
The update of weights $w_{j,t}$ is broken into two cases. From the beginning, the network is trained using \CE{} for a set number of epochs (the warm-up period $E_{\text{warmup}}$) to reach the apex phase. That is,
\[w_{j,t}=1,\gap\text{if }t\leq E_{\text{warmup}}.\]
After the warm-up period, the network throws away instances accumulating a set number (the wait period $W$) of consecutive negative margins. That is,
\begin{equation}\label{w-neg-removed}
w_{j,t+1}=\indic{\max_{i=0,1,\dots,W-1}\{m_{\theta_{t+1-i}}(x_j,y_j)\}\geq0}\indic{w_{j,t}\neq0},\gap\text{if }t\geq E_{\text{warmup}},
\end{equation}
where $m_{\theta_t}(x_j,y_j)$ is the margin of instance $j$ at epoch $t$. In other words, an instance is retained only if the previous $W$ margins are not all negative \emph{and} if the instance is still in the training set. The indicator $\indic{w_{j,t}\neq0}$ ensures that once an instance is removed, the removal is permanent.

To put it another way, if the network consistently predicts a label at odds with the observed label for an instance, then this instance is deemed noisy, and we give up learning on it (and we should). Viewed alternatively, \name{} continues to learn on clean instances and early-stops its learning on (suspected) noisy instances, filtering them out along the training process.


\begin{algorithm}[tb]
   \caption{(Mini-batch) \name}
   \label{alg-neg-removed}
\begin{algorithmic}[1]
   \STATE {\bfseries Parameters:} warm-up period \code{WARM}, wait period \code{WAIT}
   \STATE {\bfseries Input:} discriminator $f$
   \STATE {\bfseries Output:} updated discriminator $f$
   \STATE $W[:,0]\leftarrow 1$\hfill\COMMENT{Initialize weights to 1}
   \STATE $H[:,0]\leftarrow+\infty$\hfill\COMMENT{Initialize margins to $+\infty$}
   \REPEAT
   \STATE Fetch a mini-batch $(x,y)=\{(x_j,y_j)\}\jtom$ with index set $M\subset[n]$ from current epoch $e$
   \STATE $w\leftarrow W[M:e-1]$
   \STATE $logits\leftarrow$ forward$(x)$
   \STATE $costs\leftarrow$ cross entropy$(logits,y)$
   \IF{$e\leq$ \code{WARM}}
   \STATE $loss\leftarrow$ mean$(costs)$
   \STATE $margins\leftarrow H[M:e-1]$
   \ELSE
   \STATE $w\leftarrow w/$sum$(w)$\hfill\COMMENT{Normalize $w$ to sum up to 1}\label{line-normal}
   \STATE $loss\leftarrow$ sum$(w*costs)$\hfill\COMMENT{Weighted mean of $costs$}
   \STATE $margins\leftarrow logits*y$\label{line-cal-margin}
   \STATE $w\leftarrow$ update weights$(w,margins)$\label{line-update-w}
   \STATE $w[\max H[M,e-\code{WAIT}+1:e]<0]\leftarrow0$\hfill\COMMENT{Remove instances}\label{line-remove}
   \ENDIF
   \STATE update $f$ on $loss$ via SGD
   \STATE $W[M,e],H[M,e]\leftarrow w,margins$\hfill\COMMENT{Save weights and margins to history}
   \UNTIL end of training
\end{algorithmic}
\end{algorithm}

\begin{algorithm}[tb]
   \caption{Update Weights}
   \label{alg-update-w}
\begin{algorithmic}[1]
   \STATE {\bfseries Input:} weights $w$, $margins$
   \STATE {\bfseries Output:} $w$
   \STATE $w[w\neq0]\leftarrow 1$
\end{algorithmic}
\end{algorithm}


Algorithm~\ref{alg-neg-removed} illustrates this method. The weight history $W$ is a matrix where $W[i,e]$ stores the weight of instance $i$ at epoch $e$; similarly for the margin history $H$. The weight of each instance takes on values from $\{0,1\}$ throughout the training. The way $w$ is updated (e.g., line~\ref{line-update-w}, \ref{line-remove}) ensures that once the weight of an instance hits zero, it zeros out until the end. The reset step on line~\ref{line-update-w}, using Algorithm~\ref{alg-update-w}, is necessary to restore $w$ to binary digits after weight normalization on line~\ref{line-normal}.

\begin{figure}[ht]
  \begin{subfigure}[t]{\imagewidth}
    \includegraphics[width=\subwidth]{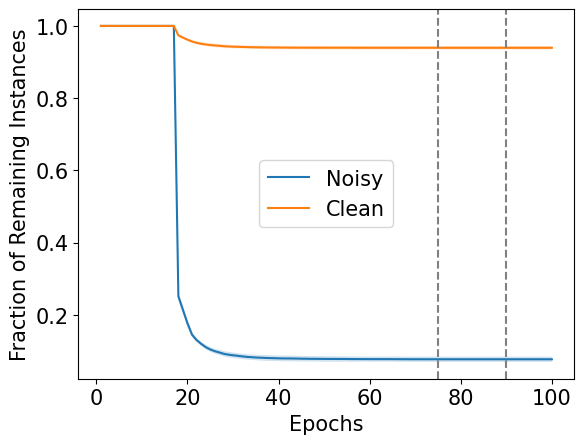}
    \caption{}
    \label{05}
  \end{subfigure}\hfill
  \begin{subfigure}[t]{\imagewidth}
    \includegraphics[width=\subwidth]{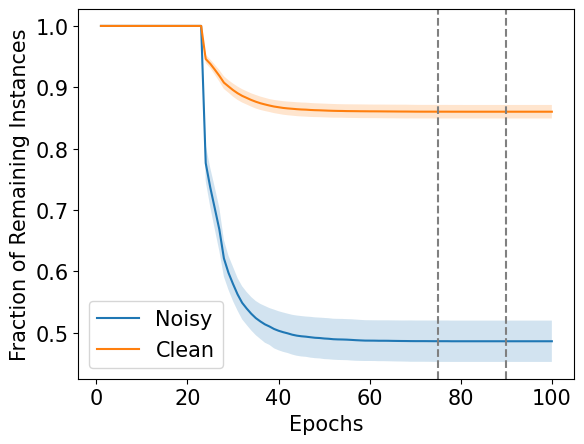}
    \caption{}
    \label{11}
  \end{subfigure}
  \caption{\name+ fraction of instances with noisy (blue) and clean (orange) labels on CIFAR-10 (a) (truck, car) with 20\% symmetric noise, and (b) (cat, dog) with 40\% asymmetric noise.}
  \label{03-all}
\end{figure}

This procedure of cleaning away noisy instances proves effective empirically on CIFAR-10. Figure~\ref{03-all} shows the fraction of remaining clean and noisy instances on different datasets. As seen from Figure~\ref{05}, the fraction of remaining noisy instances decreases drastically after the warm-up period, and plateaus at around 10\% in just a few epochs. On the other hand, the clean instances are hardly removed. For the more challenging dataset in Figure~\ref{11}, eventually half of the noisy instances are filtered out, whereas over 85\% of the clean instances are retained.

\subsection{Adaptive Weighting}\label{subsec-adapt-weight}

%

A key observation is that the method as illustrated in Algorithm~\ref{alg-neg-removed} may be thought of as training using a weighted \CE, where the weight of an instance comes from $\{0,1\}$, with the property that once zeroed out, the weight is permanently zero. Instead of binary weights, a more sophisticated weighting scheme may help facilitate the learning process. In particular, we wish to upweight difficult instances, which prove helpful in fine-tuning the decision boundary, while downweighting noisy instances. Note that easy instances, useful for learning a coarse representation but not finer features, are implicitly downweighted compared with difficult instances.

An effective way to differentiate easy, difficult or noisy instances during training is via classification margins. When the network enters the apex phase, an easy instance typically has a large margin (due to its small loss), a difficult one has a small (but positive) margin (since the network is still able to classify it correctly), and a noisy one has a smaller, or even negative, margin. This means that we wish to upweight instances with moderate margins and downweight those with small margins.

\begin{figure}[ht]
  \begin{subfigure}[t]{\imagewidth}
    \includegraphics[width=\subwidth]{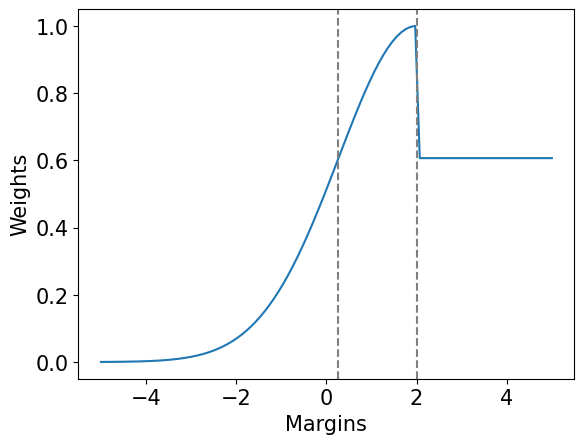}
    \caption{}
    \label{36}
  \end{subfigure}\hfill
  \begin{subfigure}[t]{\imagewidth}
    \includegraphics[width=0.95\subwidth]{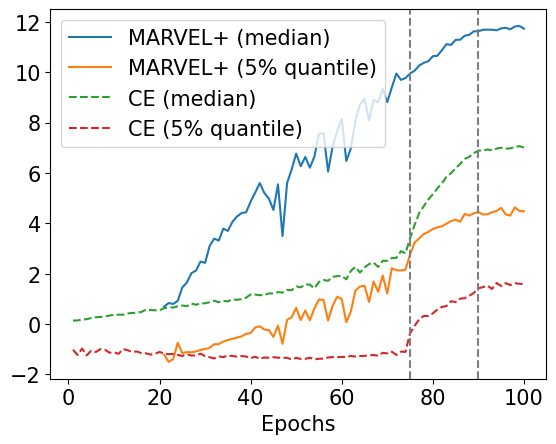}
    \caption{}
    \label{37}
  \end{subfigure}
  \caption{(a) Weights with $\hat{\mu}=2$ and $\hat{\sigma}^2=3$ (dashed),
and (b) \name+ and \CE{} 5\% quantile and median of margins on CIFAR-10 (cat, dog) with (30\%, 10\%) asymmetric noise.}
\end{figure}

Hence, for each instance, we propose to compute its weight using the following formula,
\[w(m)=\begin{cases}
\exp\brac*{-\dfrac{(m-\hat{\mu})^2}{2\hat{\sigma}^2}},&m\leq\hat{\mu};\\
e^{-1/2},&\text{o.w.},
\end{cases}\]
where $m$ is its margin, and $\hat{\mu}$ and $\hat{\sigma}^2$ are the (full-batch) computed median (not mean) and variance of the margins at the current epoch, respectively. Note that $e^{-1/2}$ is the weight of an instance with margin $\hat{\mu}-\hat{\sigma}$, i.e., one standard deviation below the median, which we call the benchmark weight.
Figure~\ref{36} shows a plot of the weights as a function of the margins.

To apply the adaptive weights, we update $w_{j,t}$ after the warm-up period using, instead of~\eqref{w-neg-removed},
\[w_{j,t+1}=\begin{cases}
\tilde{w}_{j,t+1}\cdot\exp\brac*{-\dfrac{(m_{\theta_t}(x_j,y_j)-\hat{\mu}_t)^2}{2\hat{\sigma}_t^2}},&\text{if }m_{\theta_{t-1}}(x_j,y_j)\leq\hat{\mu}_t;\\
\tilde{w}_{j,t+1}\cdot e^{-1/2},&\text{o.w.},
\end{cases}\]
where $\hat{\mu}_t$ and $\hat{\sigma}_t^2$ are the (full-batch) median and variance estimates of the margins at epoch $t$, respectively, and $\tilde{w}_{j,t+1}$ is the expression in~\eqref{w-neg-removed}. In terms of pseudo-code, the only modification to Algorithm~\ref{alg-neg-removed} is the use of Algorithm~\ref{alg-left-adaptive} to update weights on line~\ref{line-update-w}.

\begin{algorithm}[tb]
   \caption{Adaptive Weights}
   \label{alg-left-adaptive}
\begin{algorithmic}[1]
   \STATE {\bfseries Input:} weights $w$, $margins$
   \STATE {\bfseries Output:} updated weights $w$
   \STATE $w[w\neq0]\leftarrow 1$
   \STATE $median\leftarrow$ median$(margins)$
   \STATE $var\leftarrow$ variance$(margins)$
   \STATE $w[margins[i]\leq median]\leftarrow\exp[-(margins-median)^2/(2*var)]$
   \STATE $w[margins[i]>median]\leftarrow e^{-1/2}$
\end{algorithmic}
\end{algorithm}

\begin{figure}[ht]
  \begin{subfigure}[t]{\imagewidth}
    \includegraphics[width=\subwidth]{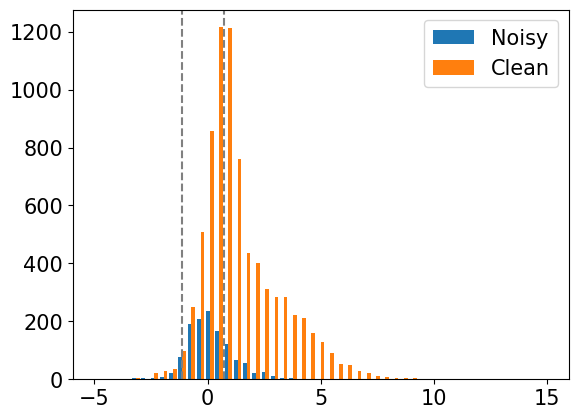}
    \caption{}
    \label{14}
  \end{subfigure}\hfill
  \begin{subfigure}[t]{\imagewidth}
    \includegraphics[width=\subwidth]{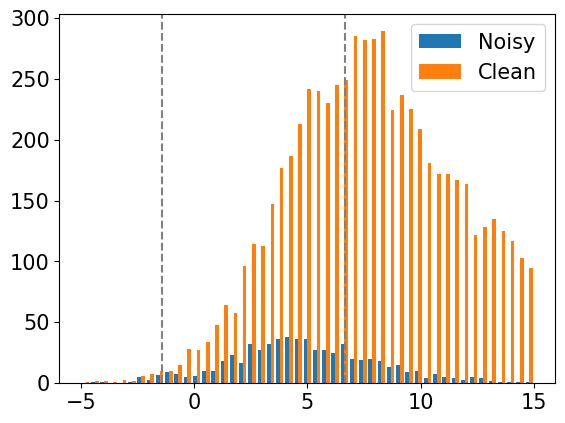}
    \caption{}
    \label{16}
  \end{subfigure}
  \caption{\name+ margin distribution at epoch 25 (a) and 75 (b) with median and standard deviation (dashed) on CIFAR-10 (cat, dog) with (30\%, 10\%) asymmetric noise.}
  \label{05-all}
\end{figure}

The threshold is set to the median because it is a robust estimate of the center. The margins roughly follow a normal distribution, as shown in Figure~\ref{05-all}. Thus most margins are around the center, of which most belong to difficult instances (with moderate margins). In this way, we maximize the number of difficult instances to be upweighted. In each mini-batch, half of the instances are set to the benchmark weight, whereas the rest are either upweighted or downweighted depending on their margins.

Note that as the removal of noisy instances kicks in, the margins of all instances (clean or noisy) shift to the right considerably. Simply testing whether the margin is positive (i.e., whether the instance is correctly classified) is not sufficient to determine whether it is clean or noisy. In Figure~\ref{37}, which displays the median and the 5\% quantile of the margins during training, we observe that the margins under training with \name{} soar up rapidly. As shown from the 5\% quantile curve, the vast majority of the margins are positive after epoch 50.

\subsection{Multi-Class Extension}

\name{} can be extended naturally to multi-class classification. For a $k$-class problem, where $k>2$, let $f:\mathcal{X}\to\real^k$ be a discriminator function. Again, the output $f(x)\in\real^k$ is called logit, the predicted class is $\argmax_j f(x)_j$, and the predicted probability of class $i$ is
\[p_i(x)=\dfrac{e^{f(x)_i}}{\sum\jtok e^{f(x)_j}},\gap i=1,2,\dots,k.\]
A natural way to define the classification margin is
\begin{equation}\label{multi-margin}
m(x,y)=f(x)_y-\max_{j\neq y}f(x)_j,
\end{equation}
so that the property is preserved that a correctly classified instance has a positive margin, while a misclassified instance has a negative margin. Algorithm~\ref{alg-neg-removed} is still applicable, but with a more generalized way, as outlined in~\eqref{multi-margin}, of calculating the margins on line~\ref{line-cal-margin}.

\subsection{Warm-up and Wait Periods}


Ideally, the warm-up period is set to the number of epochs the network takes to finish the learning phase and proceed into the apex phase. In other words, the optimal warm-up period brings the memorization ratio to the minimal level. However, this duration is unknown without the oracle noise information.

\begin{figure}[ht]
  \begin{subfigure}[t]{\imagewidth}
    \includegraphics[width=\subwidth]{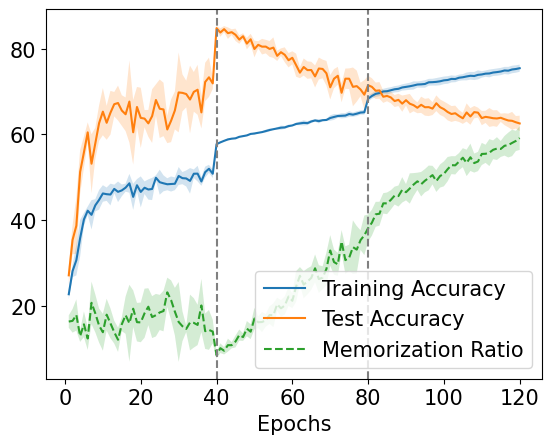}
    \caption{The warm-up period is set to 30 using either method.}
    \label{41}
  \end{subfigure}\hfill
  \begin{subfigure}[t]{\imagewidth}
    \includegraphics[width=\subwidth]{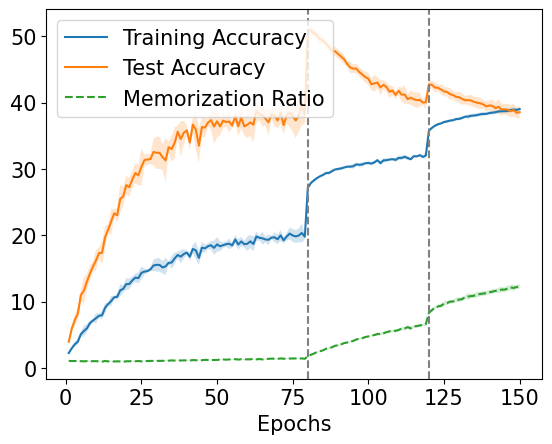}
    \caption{The warm-up period is set to 50 (orange, with clean validation set) or 40 (blue, otherwise).}
    \label{42}
  \end{subfigure}
  \caption{\CE{} on (a) CIFAR-10 with 40\% asymmetric noise, and (b) CIFAR-100 with 60\% symmetric noise.}
  \label{07-all}
\end{figure}

In practice, we set the warm-up period using the following strategy. First, we train the network using \CE. If there is a clean validation set available, we set the warm-up period to the point where the validation accuracy roughly attains its peak; otherwise, we pick the point where the rate of increase of the training accuracy starts to plateau. At this point, presumably the network has learned most of the information from the clean instances. With the gradients of clean instances shrinking to zero, the network slows down its learning, and this means that the (fast) learning phase is complete. Note that the warm-up period is dependent on the training set, as shown in Figure~\ref{07-all}, which displays the relevant curves for setting the warm-up period.

Conceptually, the warm-up period does not impact significantly the network performance. As long as the network finishes warming up and starts identifying and removing noisy instances before memorization occurs, we expect it to perform well. For instance, from Figure~\ref{41}, it seems that the warm-up period can be set to anywhere from 30 to 40 epochs, because the memorization ratio is the lowest during these epochs. Indeed, it is shown in Section~\ref{subsec-sens-anal} that a warm-up period of 30 or 40 epochs works equally well.



The wait period depends on the inherent difficulty of the dataset and the noise level. In summary, when the dataset is easy and the noise level is low, a shorter wait period works better; otherwise, a longer wait period is beneficial. In practice, we identify the optimal wait period via cross validation. We discuss in detail the empirical results on the wait period in Section~\ref{subsec-sens-anal}.


\section{Experiments}

\subsection{Binary Datasets}\label{subsec-datasets}

\paragraph{Datasets} We consider three benchmark datasets in computer vision: FASHION-MNIST, CIFAR-10 and CIFAR-100, and first focus on binary sub-datasets. In particular, we consider (boot, sneaker) in FASHION-MNIST, (truck, car) and (cat, dog) in CIFAR-10, and (aquatic, fish) in CIFAR-100; for the last one, the two labels refer to superclasses ``aquatic mammals'' and ``fish''.

\paragraph{Label noise} For symmetric noise, we randomly flip a fraction of labels in each class to the other class according to the noise level. For asymmetric noise, the fractions of labels to flip in each class are different. For instance, for (truck, car) dataset with $(30\%,10\%)$ noise, 30\% truck images are randomly flipped to cars, while 10\% car images are randomly flipped to trucks. We refer the reader to Section~\ref{subsec-exper-binary} for more details on the experimental setup.


\paragraph{Baselines} We consider the following baseline methods: (a) \textbf{CE}: training with cross entropy loss; (b) \textbf{GCE} \citep{zhang2018}: training with a robust loss that combines cross entropy and MAE; (c) \textbf{SCE} \citep{wang2019}: training with a combination of two symmetric cross entropy terms; (d) \textbf{NPCL} \citep{lyu2020}: training with a tighter upper bound of $0-1$ loss.

\begin{table}[ht]
\caption{Mean $\pm$ standard deviation (over 10 runs) of the test accuracy with symmetric label noise.}
\label{table-results-binary-sym}
\tableformat
\begin{tabular}{ccccccccc}
\toprule
Dataset &   \multicolumn{4}{c}{\textsc{Boot}}                                                                                    &   \multicolumn{4}{c}{\textsc{Truck}} \\
Noise     &                        10\% &                       20\% &                       30\% &                       40\%      &                        10\% &                       20\% &                       30\% &                       40\% \\
\midrule
CE         &                 $92.0\pm0.8$ & $84.5\pm1.3$ & $74.9\pm2.2$ & $68.3\pm4.3$               &                 $89.8\pm0.5$ & $81.2\pm1.0$ & $70.3\pm1.3$ & $60.4\pm1.5$ \\
GCE       &                 ${\second{97.1}}\pm0.2$ & $96.0\pm0.7$ & $\best{96.1}\pm0.3$ & $94.3\pm0.5$      &                  $94.8\pm0.3$ & $93.3\pm0.4$ & $89.3\pm0.8$ & $75.0\pm2.0$ \\
SCE        &                 $\best{97.2}\pm0.2$ & $96.4\pm0.5$ & $94.1\pm0.7$ & $89.0\pm2.7$       &                  $94.8\pm0.3$ & $93.5\pm0.4$ & $89.1\pm1.1$ & $77.6\pm2.4$ \\
NPCL      &                 ${\second{97.1}}\pm0.2$ & ${\second{96.8}}\pm0.2$ & ${\second{96.0}}\pm0.3$ & $93.1\pm0.6$         &                  ${\second{95.2}}\pm0.2$ & $93.8\pm0.6$ & ${\second{90.0}}\pm0.8$ & $71.3\pm1.1$ \\
\name     &                 ${\second{97.1}}\pm0.2$ & ${\second{96.8}}\pm0.2$ & ${\second{96.0}}\pm0.3$ & $\best{94.4}\pm0.4$       &                $95.0\pm0.4$ & $\best{93.9}\pm0.3$ & $89.9\pm0.5$ & $\best{79.6}\pm2.0$ \\
\name+                    &                 $97.0\pm0.2$ & $\best{96.9}\pm0.3$ & $95.9\pm0.3$ & $\best{94.4}\pm0.5$       &                  $\best{95.3}\pm0.3$ & $\best{93.9}\pm0.4$ & $\best{90.3}\pm0.5$ & ${\second{79.3}}\pm1.8$\\
\midrule
\midrule
Dataset &   \multicolumn{4}{c}{\textsc{Cat}}                                                                                    &   \multicolumn{4}{c}{\textsc{Aquatic}} \\
Noise     &                        10\% &                       20\% &                       30\% &                       40\%               &                         10\% &                       20\% &                       30\% &                       40\% \\
\midrule
CE         &                 $82.0\pm0.5$ & $73.4\pm0.8$ & $64.7\pm1.5$ & $56.6\pm1.1$            &                  $78.6\pm1.0$ & $71.4\pm1.2$ & $63.4\pm1.9$ & $59.3\pm2.0$ \\
GCE       &               $86.1\pm0.7$ & $\best{83.6}\pm0.7$ & $\best{76.7}\pm1.4$ & $63.0\pm1.8$          &                  $82.6\pm0.7$ & $80.1\pm1.5$ & $75.6\pm1.8$ & $67.5\pm2.2$ \\
SCE        &                $86.7\pm0.5$ & $83.3\pm0.8$ & $75.0\pm1.6$ & $63.2\pm2.1$              &                  ${\second{83.1}}\pm0.7$ & $80.1\pm1.4$ & $75.9\pm2.2$ & $67.9\pm1.9$ \\
NPCL      &                 $86.0\pm0.4$ & $82.9\pm0.7$ & $74.9\pm1.9$ & $60.4\pm2.9$          &                  $82.7\pm0.8$ & $79.9\pm1.1$ & $75.2\pm1.7$ & $65.4\pm2.5$ \\
\name     &                 ${\second{87.1}}\pm0.4$ &$ 83.3\pm0.8$ & $76.2\pm1.3$ & ${\second{65.6}}\pm1.1$            &                   $83.0\pm0.9$ & ${\second{80.3}}\pm0.6$ & ${\second{76.0}}\pm1.9$ & ${\second{69.9}}\pm1.9$ \\
\name+                    &                 $\best{87.7}\pm0.3$ & $\best{83.6}\pm0.9$ & ${\second{76.6}}\pm1.4$ & $\best{66.5}\pm2.0$             &                  $\best{84.0}\pm0.7$ & $\best{80.9}\pm0.4$ & $\best{76.8}\pm1.7$ & $\best{70.9}\pm1.5$ \\
\bottomrule
\end{tabular}

\end{table}

\begin{table}[ht]
\caption{Mean $\pm$ standard deviation (over 10 runs) of the test accuracy with asymmetric label noise.}
\label{table-results-binary-asym}
\tableformat
\begin{tabular}{ccccccccc}
\toprule
Dataset &   \multicolumn{4}{c}{\textsc{Boot}}                                                                                    &   \multicolumn{4}{c}{\textsc{Truck}} \\
Noise     &                        (20\%,0\%) &                       (30\%,10\%) &                       (40\%,0\%) &                       (40\%,10\%)      &                         (20\%,0\%) &                       (30\%,10\%) &                       (40\%,0\%) &                       (40\%,10\%) \\
\midrule
CE         &                 $93.1\pm1.7$ & $82.4\pm1.8$ & $84.3\pm2.8$ & $78.0\pm1.7$              &                  $89.1\pm0.8$ & $79.4\pm0.9$ & $78.0\pm1.1$ & $74.4\pm0.8$ \\
GCE       &                 $96.6\pm0.4$ & $96.0\pm0.3$ & $88.2\pm1.9$ & $87.9\pm1.9$      &                  $94.1\pm0.4$ & $91.3\pm0.8$ & $78.0\pm1.0$ & $81.7\pm2.2$ \\
SCE        &                $96.7\pm0.3$ & $95.0\pm0.8$ & $91.2\pm1.3$ & $90.0\pm2.0$      &                  $94.2\pm0.3$ & $91.6\pm0.7$ & $82.0\pm2.1$ & $83.0\pm2.2$ \\
NPCL      &                 $96.8\pm0.3$ & $96.1\pm0.3$ & $94.0\pm1.1$ & $93.6\pm1.4$         &                  $94.7\pm0.4$ & $92.1\pm0.8$ & $80.2\pm1.8$ & $81.2\pm4.0$ \\
\name     &                 $\best{97.0}\pm0.2$ & $\best{96.3}\pm0.2$ & ${\second{95.2}}\pm0.3$ & ${\second{94.2}}\pm0.6$       &                ${\second{95.4}}\pm0.2$ & ${\second{93.4}}\pm0.5$ & ${\second{92.4}}\pm0.6$ & ${\second{88.7}}\pm1.3$ \\
\name+                    &                 $\best{97.0}\pm0.1$ & $\best{96.3}\pm0.3$ & $\best{95.5}\pm0.2$ & $\best{95.0}\pm0.3$       &                  $\best{95.6}\pm0.2$ & $\best{93.7}\pm0.4$ & $\best{92.9}\pm0.5$ & $\best{89.8}\pm1.1$\\
\midrule
\midrule
Dataset &   \multicolumn{4}{c}{\textsc{Cat}}                                                                                    &   \multicolumn{4}{c}{\textsc{Aquatic}} \\
Noise     &                        (20\%,0\%) &                       (30\%,10\%) &                       (40\%,0\%) &                       (40\%,10\%)               &                         (20\%,0\%) &                       (30\%,10\%) &                       (40\%,0\%) &                       (40\%,10\%) \\
\midrule
CE         &                 $82.1\pm0.8$ & $73.2\pm1.2$ & $72.7\pm1.2$ & $68.1\pm1.3$            &                  $79.0\pm1.0$ & $71.5\pm1.2$ & $71.3\pm1.5$ & $68.0\pm1.2$ \\
GCE       &                 $83.7\pm1.3$ & $78.3\pm0.9$ & $72.5\pm1.3$ & $70.3\pm1.3$          &                  $82.0\pm0.9$ & $78.3\pm1.3$ & $70.7\pm1.2$ & $70.6\pm2.5$ \\
SCE        &                 $84.8\pm0.9$ & $78.4\pm1.2$ & $72.8\pm1.7$ & $66.5\pm1.4$              &                  $81.7\pm0.6$ & $77.3\pm1.2$ & $69.4\pm1.1$ & $70.3\pm1.9$ \\
NPCL      &                 $83.1\pm0.7$ & $76.2\pm1.8$ & $62.0\pm0.9$ & $61.8\pm1.6$        &                  $81.9\pm0.9$ & $78.5\pm1.4$ & $67.5\pm2.4$ & $68.6\pm2.7$ \\
\name     &                 ${\second{86.9}}\pm1.5$ & ${\second{81.3}}\pm1.1$ & ${\second{75.5}}\pm1.7$ & ${\second{70.7}}\pm2.3$            &                   ${\second{82.6}}\pm0.8$ & ${\second{79.2}}\pm1.0$ & ${\second{71.4}}\pm4.7$ & ${\second{71.4}}\pm2.8$ \\
\name+                    &                  $\best{87.3}\pm0.5$ & $\best{82.3}\pm1.1$ & $\best{77.5}\pm2.3$ & $\best{72.6}\pm1.4$             &                 $\best{83.6}\pm0.5$ & $\best{80.1}\pm0.9$ & $\best{74.5}\pm1.8$ & $\best{74.0}\pm2.0$ \\
\bottomrule
\end{tabular}

\end{table}

\paragraph{Results and Discussions} Table~\ref{table-results-binary-sym} and \ref{table-results-binary-asym} evaluate the performance of all methods on different binary datasets with various noise levels, from which we observe that \name{} and \name+ consistently outperform other baselines; the margin is significantly larger under asymmetric noise. Furthermore, the standard deviation of \name{} is consistently smaller than other robust methods, albeit slightly less so for more challenging datasets (cat, dog) and (aquatic, fish).


Note that NPCL's performance degrades rapidly when the noise level is asymmetric and highly skewed. We hypothesize that the low-loss instances NPCL selects in each mini-batch are mainly instances from the less noisy class, because the noisier class contains more instances with high losses. This results in an imbalance in the classes, which likely contributes to the degradation of performance, in the same way as the more diversified SPLD \citep{jiang2014} outperforms vanilla SPL \citep{kumar2010}.

\begin{figure}[ht]
  \begin{subfigure}[t]{\imagewidth}
    \includegraphics[width=\subwidth]{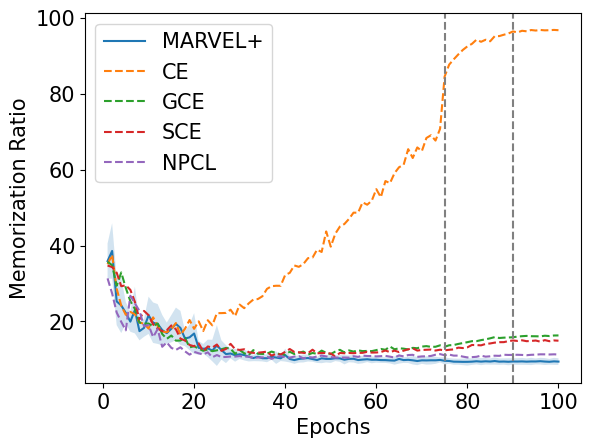}
    \caption{}
    \label{19}
  \end{subfigure}\hfill
  \begin{subfigure}[t]{\imagewidth}
    \includegraphics[width=\subwidth]{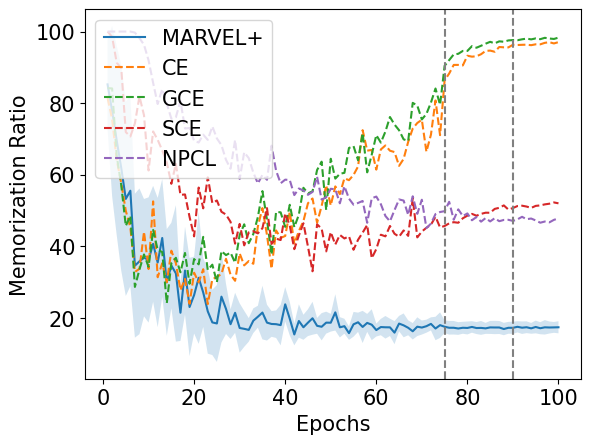}
    \caption{}
    \label{21}
  \end{subfigure}
   \begin{subfigure}[t]{\imagewidth}
    \includegraphics[width=\subwidth]{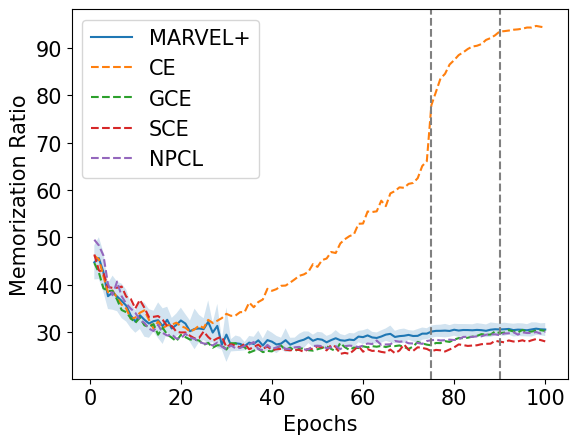}
    \caption{}
    \label{23}
  \end{subfigure}\hfill
   \begin{subfigure}[t]{\imagewidth}
    \includegraphics[width=\subwidth]{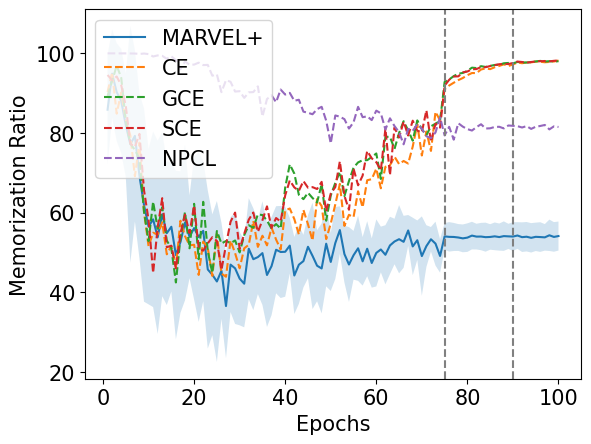}
    \caption{}
    \label{25}
  \end{subfigure}
  \caption{Memorization ratio on CIFAR-10 (truck, car) with (a) 20\% symmetric noise and (b) 40\% asymmetric noise, and (cat, dog) with (c) 20\% symmetric noise and (d) 40\% asymmetric noise.}
  \label{08-all}
\end{figure}

Figure~\ref{08-all} shows the curves of memorization ratio on different datasets. For NPCL, every second epoch is sampled to be comparable with other methods. When the noise is symmetric, all robust methods perform similarly. However, under asymmetric noise, \name+ has the advantage of refraining from memorizing the noisy labels: the memorization ratio remains at a low level throughout the training. The other robust methods (except NPCL, which struggles in the asymmetric case) gradually memorize the noisy labels, leading to a decline in the test accuracy.

\subsection{Multi-Class Datasets}

\paragraph{Datasets} We consider two benchmark datasets: CIFAR-10 and CIFAR-100.

\paragraph{Label noise} For symmetric noise, we randomly flip a fraction of labels to any \emph{other} class according to the noise level. For asymmetric noise, following \citet{pleiss2020}, we order the $c$ classes using $\{0,1,\dots,c-1\}$, and flip a fraction of the labels circularly (i.e., class 0 to class 1, class 1 to class 2, class $c-1$ to class 0, and so on). For CIFAR-10 dataset, we alternatively follow the class-dependent noise in \citep{patrini2017}, flipping truck $\rightarrow$ car, bird $\rightarrow$ airplane, deer $\rightarrow$ horse, and cat $\leftrightarrow$ dog according to the noise level. We refer the reader to Section~\ref{subsec-multi-32} and \ref{subsec-multi-34} for more details on the experimental setup.

\paragraph{Baselines} We consider the following baseline methods: (a) \textbf{CE}; (b) \textbf{SCE} \citep{wang2019}; (c) \textbf{Bootstrap} \citep{reed2015}: training with labels interpolated from the observed labels and the predicted labels; (d) \textbf{MentorNet} \citep{jiang2018}: training with a LSTM-generated data-driven curriculum; (e) \textbf{Co-teaching} \citep{han2018}: training with two cross-updating networks; (f) \textbf{D2L} \citep{ma2018}: training with a loss function adapted to subspace dimensionality; (g) \textbf{L\textsubscript{DMI}} \citep{xu2019}: training with an information-theoretic robust loss function; (h) \textbf{Data Parameters} \citep{saxena2019}: training with a learnable weighting parameter to each instance/class; (i) \textbf{DY-Bootstrap} \citep{arazo2019}: training with a mixture of beta distributions; (j) \textbf{INCV} \citep{chen2019}: training with iterative noisy cross-validation; (k) \textbf{AUM} \citep{pleiss2020}: training with area under the margin ranking. We also include the performance of an \textbf{oracle} learner, which trains using only the clean instances.

\begin{table}[ht]
\caption{Mean $\pm$ standard deviation (over 5 runs) of the test accuracy with symmetric label noise, using ResNet-32 from the official code of \citet{pleiss2020}. We run CE, SCE and Oracle, and the rest are taken from \citet{pleiss2020}, where it is noted that the MentorNet code release cannot be used with 60\% symmetric noise on CIFAR.}
\label{table-cifar-sym}
\tableformat
\begin{tabular}{ccccccccc}
\toprule
Dataset &   \multicolumn{4}{c}{\textsc{CIFAR-10}} &  \multicolumn{4}{c}{\textsc{CIFAR-100}} \\
Noise &                        20\% &                       40\% &                       60\% &                       80\% &                         20\% &                       40\% &                       60\% &                       80\% \\
\midrule
CE                                   &                 $82.3\pm0.7$ & $72.2\pm0.7$ & $51.3\pm0.9$ & $29.0\pm2.8$ &                  $58.8\pm0.4$ & $50.5\pm0.3$ & $38.6\pm0.9$ & $17.2\pm1.1$ \\
SCE                                   &                $89.5\pm0.1$ & $86.2\pm0.1$ & $78.5\pm0.3$ & $42.8\pm4.6$ &                  $55.6\pm0.5$ & $45.4\pm0.7$ & $31.8\pm1.1$ & $14.2\pm2.4$ \\
Bootstrap         &                 $77.6 \pm 0.2$ &             $62.6 \pm 0.4$ &             $48.0 \pm 0.2$ &             $31.2 \pm 0.5$ &                  $51.4 \pm 0.2$ &             $41.1 \pm 0.2$ &             $29.7 \pm 0.2$ &             $10.2 \pm 0.8$ \\
MentorNet       &                 $86.7 \pm 0.1$ &             $81.9 \pm 0.2$ &                       --- &             $34.0 \pm 2.5$ &                  $64.2 \pm 0.3$ &             $57.5 \pm 0.2$ &                       --- &             $24.3 \pm 0.7$ \\
Co-Teaching      &                 $88.8 \pm 0.1$ &             $86.5 \pm 0.1$ &             $80.7 \pm 0.1$ &             $19.3 \pm 0.6$ &                  $64.1 \pm 0.1$ &             $60.2 \pm 0.2$ &             $48.0 \pm 0.3$ &             $10.9 \pm 0.7$ \\
D2L                      &                 $87.7 \pm 0.2$ &             $84.4 \pm 0.3$ &             $72.7 \pm 0.6$ &                   Diverged &                  $54.0 \pm 1.0$ &             $29.7 \pm 1.8$ &                   Diverged &                   Diverged \\
L\textsubscript{DMI}         &                 $85.9 \pm 0.1$ &             $79.6 \pm 0.8$ &             $65.1 \pm 1.2$ &             $32.8 \pm 1.4$ &                        Diverged &                   Diverged &                   Diverged &                   Diverged \\
Data Params          &                 $82.1 \pm 0.2$ &             $70.8 \pm 0.9$ &             $49.3 \pm 0.7$ &             $18.9 \pm 0.3$ &                  $56.3 \pm 1.4$ &             $46.1 \pm 0.4$ &             $32.8 \pm 2.3$ &             $11.9 \pm 1.2$ \\
DY-Bootstrap  &                 $79.4 \pm 0.1$ &             $68.8 \pm 1.4$ &             $56.4 \pm 1.7$ &                   Diverged &                  $53.0 \pm 0.4$ &             $43.0 \pm 0.3$ &             $36.6 \pm 0.5$ &             $12.8 \pm 0.5$ \\
INCV                  &                 $89.5 \pm 0.1$ &             $86.8 \pm 0.1$ &             $81.1 \pm 0.3$ &  ${\second{53.3}} \pm 1.9$ &                  $58.6 \pm 0.5$ &             $55.4 \pm 0.2$ &             $43.7 \pm 0.3$ &             $23.7 \pm 0.6$ \\
AUM                                        &       $90.2 \pm 0.0$ &  $87.5 \pm 0.1$ &  $82.1 \pm 0.0$ &  $\best{54.4} \pm 1.6$ &       $65.5 \pm 0.2$ &  $61.3 \pm 0.1$ &  $53.0 \pm 0.5$ &  $\best{31.7} \pm 0.7$ \\
\name                                &                 $\best{91.1}\pm0.2$ & $\best{88.4}\pm0.2$ & $\best{83.1}\pm0.1$ & $52.1\pm4.2$ &                  ${\second{67.5}}\pm0.2$ & $\best{63.8}\pm0.2$ & ${\second{56.8}}\pm0.3$ & ${\second{31.3}}\pm2.2$ \\
\name+                               &                 ${\second{91.0}}\pm0.2$ & $\best{88.4}\pm0.2$ & ${\second{83.0}}\pm0.2$ & $48.4\pm3.9$ &                  $\best{67.7}\pm0.1$ & ${\second{63.6}}\pm0.3$ & $\best{56.9}\pm0.7$ & $30.9\pm1.8$ \\
\midrule
Oracle                                  &                 $91.3\pm0.3$ & $90.0\pm0.2$ & $88.0\pm0.4$ & $83.5\pm1.3$ &                  $67.6\pm0.2$ & $64.8\pm0.5$ & $60.8\pm0.6$ & $51.3\pm0.4$ \\
\bottomrule
\end{tabular}

\end{table}

\begin{table}[ht]
\caption{Mean $\pm$ standard deviation (over 5 runs) of the test accuracy with asymmetric label noise.}
\label{table-cifar-asym}
\tableformat
\begin{tabular}{ccccc}
\toprule
Dataset &  \multicolumn{2}{c}{\textsc{CIFAR-10}} &      \multicolumn{2}{c}{\textsc{CIFAR-100}} \\
Noise &                       20\% &                       40\% &                       20\% &                       40\% \\
\midrule
CE                                   &                 $81.9\pm0.9$ & $62.5\pm1.8$ &                              $60.5\pm0.2$ & $44.8\pm0.8$             \\
SCE                                   &                 $88.8\pm0.2$ & $76.7\pm1.7$ &                              $56.5\pm0.5$ & $43.5\pm0.5$            \\
Bootstrap         &            $76.2 \pm 0.2$ &             $55.0 \pm 0.6$ &             $53.4 \pm 0.3$ &             $38.7 \pm 0.3$ \\
D2L                      &             $88.6 \pm 0.2$ &             $76.4 \pm 1.5$ &            $43.6 \pm 0.7$ &             $16.9 \pm 1.2$ \\
L\textsubscript{DMI}         &             $86.7 \pm 0.8$ &  $84.0 \pm 2.1$ &                   Diverged &                   Diverged \\
Data Params          &             $82.1 \pm 0.2$ &             $55.5 \pm 0.7$ &             $56.2 \pm 0.5$ &             $39.0 \pm 0.4$ \\
DY-Bootstrap &             $77.9 \pm 0.1$ &             $59.4 \pm 0.6$ &             $53.2 \pm 0.0$ &             $37.9 \pm 0.0$ \\
INCV                  &             $88.3 \pm 0.1$ &             $79.8 \pm 0.4$ &             $56.8 \pm 0.1$ &  $44.4 \pm 0.7$ \\
AUM                                     &     $89.7 \pm 0.1$ &             $58.7 \pm 0.2$ &       $59.7 \pm 0.2$ &             $40.2 \pm 0.1$ \\
\name                                &                 $\best{91.5}\pm0.2$ & $\best{89.4}\pm0.3$ &                             $\best{67.8}\pm0.2$ & $\best{60.6}\pm0.3$             \\
\name+                               &                 ${\second{91.4}}\pm0.2$ & ${\second{88.7}}\pm0.2$ &                             ${\second{67.5}}\pm0.3$ & ${\second{59.5}}\pm0.4$            \\
\midrule
Oracle                                  &                 $91.5\pm0.2$ & $90.1\pm0.2$ &                              $67.5\pm0.4$ & $65.1\pm0.4$             \\
\bottomrule
\end{tabular}

\end{table}

\begin{table}[ht]
\caption{Mean $\pm$ standard deviation (over 5 runs) of the test accuracy, using ResNet-34, on CIFAR-10 with symmetric and class-dependent label noise, at the epoch with lowest validation error, except for \CE, whose final test accuracy is reported.}
\label{table-resnet34}
\tableformat
\begin{tabular}{ccccccccc}
\toprule
Type &   \multicolumn{4}{c}{\textsc{Symmetric}}                                                                                      & \multicolumn{4}{c}{\textsc{Class-Dependent}} \\
Noise     &                        20\% &                       40\% &                       60\% &                       80\%      &                         10\% &                       20\% &                       30\% &                       40\% \\
\midrule
CE         &                 $81.4\pm0.3$ & $61.2\pm1.4$ & $37.3\pm1.9$ & $29.7\pm2.5$               &                  $90.7\pm0.2$ & $86.2\pm0.4$ & $81.5\pm0.5$ & $76.1\pm0.7$ \\
SCE        &                 $90.2\pm0.2$ & $86.9\pm0.2$ & $80.9\pm0.1$ & $\best{43.9}\pm0.9$       &                  $91.5\pm0.1$ & $90.1\pm0.2$ & $88.0\pm0.1$ & $84.3\pm0.6$ \\
\name     &                 $\best{92.5}\pm0.3$ & ${\second{89.2}}\pm0.3$ & $\best{82.2}\pm2.2$ & ${\second{42.1}}\pm2.1$       &                $\best{93.7}\pm0.3$ & ${\second{92.8}}\pm0.3$ & ${\second{91.8}}\pm0.3$ & ${\second{90.6}}\pm0.3$ \\
\name+                         &            ${\second{92.2}}\pm0.2$ & $\best{89.5}\pm0.3$ & $\best{82.2}\pm1.2$ & $37.9\pm2.6$       &                 ${\second{93.6}}\pm0.2$ & $\best{93.1}\pm0.2$ & $\best{92.0}\pm0.4$ & $\best{90.8}\pm0.5$ \\
\midrule
Oracle       &                 $92.6\pm0.2$ & $91.4\pm0.4$ & $89.0\pm0.4$ & $83.2\pm1.2$      &                  $93.5\pm0.1$ & $93.3\pm0.1$ & $92.8\pm0.4$ & $92.4\pm0.1$ \\
\bottomrule
\end{tabular}

\end{table}

Table~\ref{table-cifar-sym} and \ref{table-cifar-asym} evaluate the performance of all methods on CIFAR-10 and CIFAR-100 using ResNet-32, following \citet{pleiss2020}, and Table~\ref{table-resnet34} evaluates the performance on CIFAR-10 using ResNet-34, following the class-dependent noise in \citet{patrini2017}. Again, \name{} outperforms other baselines, especially under asymmetric noise and class-dependent noise. However, under 80\% symmetric noise, \name{} does not work as effectively as expected. In this case, \name{} is still able to identify and filter out most of the noisy instances, but removing too many instances eventually hurts the network generalization. This phenomenon shows that the feature part of a noisy instance is still useful in training, especially when the noise level is high.

\begin{figure}[ht]
  \begin{subfigure}[t]{\imagewidth}
    \includegraphics[width=\subwidth]{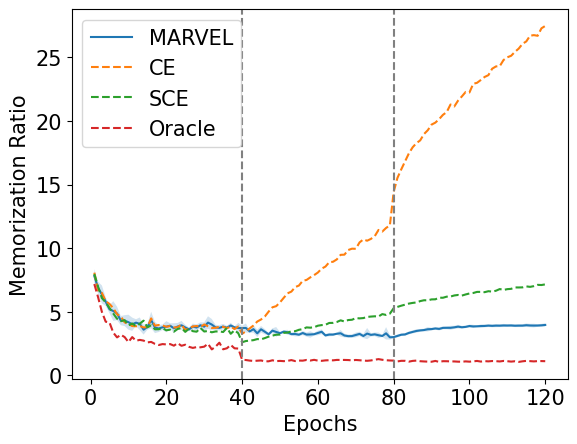}
    \caption{}
    \label{43}
  \end{subfigure}\hfill
  \begin{subfigure}[t]{\imagewidth}
    \includegraphics[width=\subwidth]{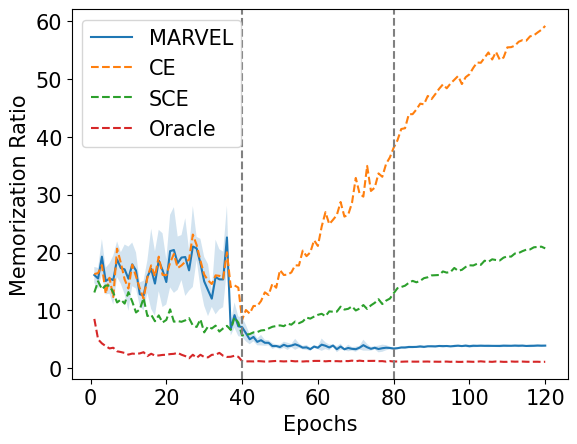}
    \caption{}
    \label{44}
  \end{subfigure}
  \begin{subfigure}[t]{\imagewidth}
    \includegraphics[width=\subwidth]{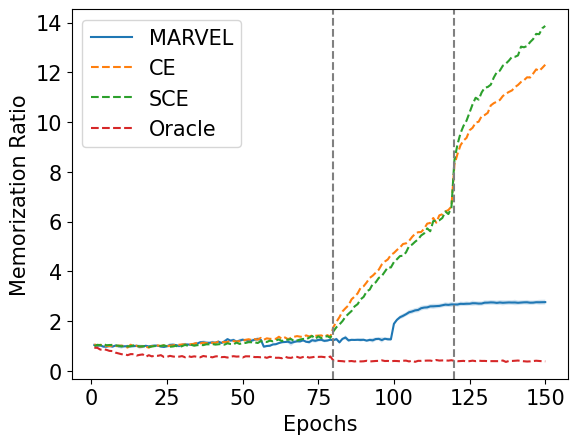}
    \caption{}
    \label{45}
  \end{subfigure}\hfill
  \begin{subfigure}[t]{\imagewidth}
    \includegraphics[width=\subwidth]{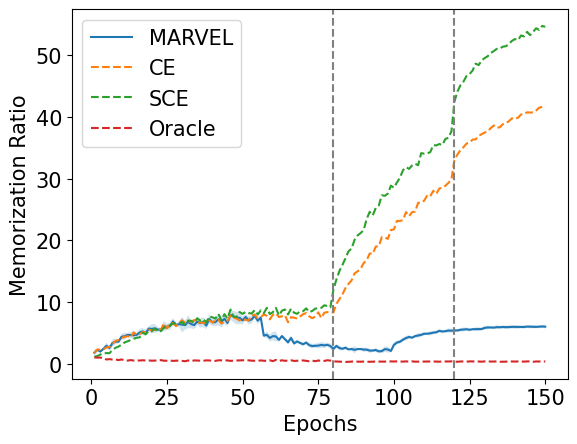}
    \caption{}
    \label{46}
  \end{subfigure}
  \caption{Memorization ratio on CIFAR-10 with (a) 40\% symmetric noise and (b) 40\% asymmetric noise, and CIFAR-100 with (c) 60\% symmetric noise and (d) 20\% asymmetric noise.}
  \label{10-all}
\end{figure}

Figure~\ref{10-all} displays the curves of memorization ratio on CIFAR-10 and CIFAR-100. Under asymmetric noise, \name{} drives down the memorization ratio as removal kicks in (after a warm-up period of 30 or 40 epochs for CIFAR-10 and CIFAR-100, respectively), while other methods succumb to the noisy labels.

\subsection{Real-World Datasets}

We test \name{} on two real-world datasets, WebVision50 and Clothing100K, and compare it with \CE{} in terms of test accuracy. WebVision \citep{li2017} contains more than 2.4 million images crawled from the Flickr website and Google Images search. Following \citet{chen2019} and \citet{pleiss2020}, we train \name{} on a subset, WebVision50, which contains the first 50 classes (126,370 images). Clothing100K is a 100K-image subset of Clothing1M \citep{xiao2015}, which contains clothing images scraped from internet and categorized into 14 classes. We refer the reader to Section~\ref{subsec-real-world} for more details on the experimental setup.

\begin{table}[ht]
\caption{Upper: Mean $\pm$ standard deviation over the last 10 test accuracy figures. Lower: Figures taken from \citet{pleiss2020}.}
\label{table-real}
\tableformat
\begin{tabular}{ccccccccc}
\toprule
Dataset &  \textsc{WebVision50} &   \textsc{Clothing100K} \\
\midrule
CE                                   &                 $73.1\pm0.4$ &  $63.9\pm0.4$ \\
\name                                   &                 $\best{74.5}\pm0.2$ &  $\best{67.2}\pm0.3$ \\
\midrule
\midrule
CE                                   &                 $78.6$ &  $64.2$ \\
Data Params            &                 $78.5$ &  $64.5$ \\
DY-Bootstrap                   &                $74.2$ &  $61.6$ \\
INCV                                   &                 $77.9$ &  $\best{66.7}$ \\
AUM                                   &                 $\best{80.2}$ &  $66.5$ \\
\bottomrule
\end{tabular}
\end{table}

Table~\ref{table-real} evaluates the performance of \name{} and \CE{} on the two datasets using ResNet-50, following \citet{pleiss2020}. \name{} achieves a performance boost of $1.4\%$ and $3.3\%$ on WebVision50 and Clothing100K, respectively, which is comparable to that of other methods considered in \citet{pleiss2020} (or even slightly superior on Clothing100K).

\subsection{Clean Label Identification}

\begin{figure}[ht]
  \begin{subfigure}[t]{\imagewidth}
    \includegraphics[width=\subwidth]{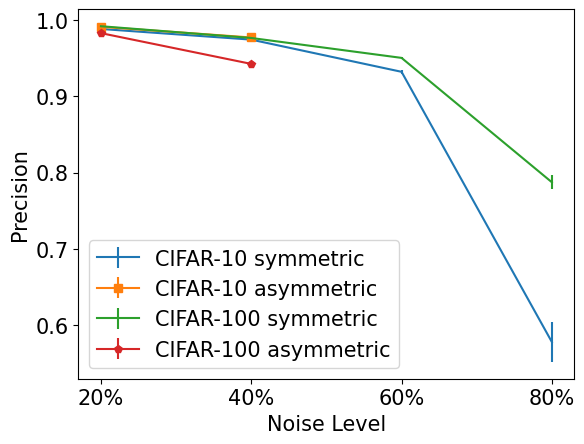}
    \caption{}
    \label{40}
  \end{subfigure}\hfill
  \begin{subfigure}[t]{\imagewidth}
    \includegraphics[width=\subwidth]{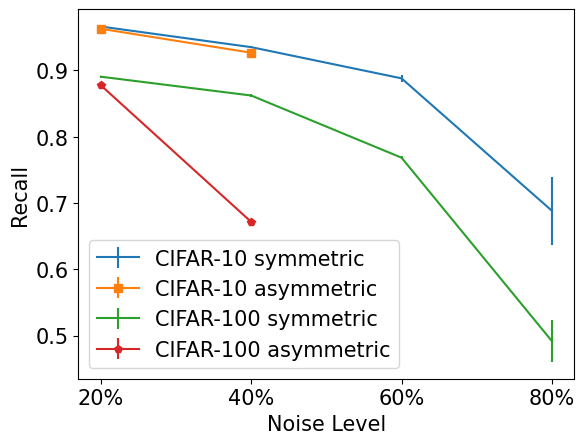}
    \caption{}
    \label{47}
  \end{subfigure}
  \caption{Precision (a) and recall (b) on CIFAR-10 and CIFAR-100. The mean (curve) and the standard deviation of 5 runs (error bar) are reported.}
  \label{11-all}
\end{figure}

Following \citet{han2018} and \citet{chen2019}, we calculate label precision and recall. Label precision is defined as the fraction of clean instances in the filtered training set $S$ at the end of training (i.e., instances not thrown away by \name), whereas label recall is defined as the fraction of clean instances in $S$ over those in the entire training set $D_{\rho}$. From Figure~\ref{11-all}, we see that \name{} is capable of identifying clean labels with high accuracy. For instance, \name{} identifies around 90\% of all the clean labels in CIFAR-10 with 60\% symmetric noise, and the filtered training set has only 7\% noisy labels.

\subsection{Sensitivity Analysis of Hyperparameters}\label{subsec-sens-anal}

\paragraph{Warm-up period} Table~\ref{table-tune-warmup} illustrates the effect of the warm-up period on the performance of \name. From the table, we see that a wide range of warm-up periods work for \name, resulting in roughly identical performance. Hence \name{} proves robust to the warm-up period.

\begin{table}[ht]
\caption{Mean $\pm$ standard deviation (over 5 runs) of the test accuracy for \name{} when the warm-up period varies. The warm-up periods used in Table~\ref{table-cifar-sym} and \ref{table-cifar-asym} are in bold.}
\label{table-tune-warmup}
\tableformat
\begin{tabular}{ccccccccccc}
\toprule
Dataset &  & \multicolumn{4}{c}{\textsc{CIFAR-10 40\% Symmetric}}                                                                                    &  & \multicolumn{4}{c}{\textsc{CIFAR-10 40\% Asymmetric}} \\
Warm-up     &  &                      20 &                       30 &                       40 &                       50      &  &                       20 &                       30 &                       40 &                       50 \\
\midrule
\name     &     &            $88.4\pm0.2$ & $\select{88.4}\pm0.2$ & $88.6\pm0.2$ & $88.3\pm0.3$        &     &           $88.9\pm0.2$ & $\select{89.4}\pm0.3$ & $89.6\pm0.1$ & $89.7\pm0.1$ \\
\midrule
\midrule
Dataset &  & \multicolumn{4}{c}{\textsc{CIFAR-100 60\% Symmetric}}                                                                                    &  & \multicolumn{4}{c}{\textsc{CIFAR-100 20\% Asymmetric}} \\
Warm-up     &  &                      20 &                       30 &                       40 &                       50      &  &                       20 &                       30 &                       40 &                       50 \\
\midrule
\name     &     &            $55.3\pm0.5$ & $56.6\pm0.2$ & $\select{56.8}\pm0.3$ & $56.7\pm0.4$            &      &             $67.4\pm0.2$ & $67.6\pm0.2$ & $\select{67.8}\pm0.2$ & $67.4\pm0.3$ \\
\bottomrule
\end{tabular}
\end{table}

\paragraph{Wait period} In Table~\ref{table-tune-wait}, we vary the wait period of \name, and see that a slight misspecification of the wait period has a negligible impact on the performance of the network, and thus \name{} proves robust to the wait period. The selected wait periods for different datasets are presented in the Appendix. In summary, we observe that the wait period depends crucially on the inherent difficulty of the dataset: a more challenging dataset benefits from a longer wait period, as displayed by CIFAR-10 and CIFAR-100 in Table~\ref{table-cifar-sym-wait} and \ref{table-cifar-asym-wait}. For a given dataset, the wait period shortens for higher or more skewed noise levels, unless the dataset is extremely noisy (e.g., (aquatic, fish) with 40\% asymmetric noise, or CIFAR-10 with 80\% symmetric noise), in which case a longer wait period works better. We conjecture that in the former case, as few truly difficult instances exist, those consistently misclassified are more likely to be noisy, and a shorter wait period prevents memorization. In the latter case, most of the instances are noisy, and a (too) short wait period permits learning only on the (relatively few) easy instances, which does not help generalization.

\begin{table}[ht]
\caption{Mean $\pm$ standard deviation (over 5 runs) of the test accuracy for \name{} when the wait period varies. The wait periods selected in Table~\ref{table-cifar-sym} and \ref{table-cifar-asym} are in bold.}
\label{table-tune-wait}
\tableformat
\begin{tabular}{ccccccccc}
\toprule
Dataset &   \multicolumn{4}{c}{\textsc{CIFAR-10 40\% Symmetric}}                                                                                      & \multicolumn{4}{c}{\textsc{CIFAR-10 40\% Asymmetric}} \\
Wait     &                        6 &                       8 &                       10 &                       12      &                         4 &                       6 &                       8 &                       10 \\
\midrule
\name     &                 $88.4\pm0.2$ & $\select{88.4}\pm0.2$ & $88.2\pm0.2$ & $88.0\pm0.1$        &                $88.9\pm0.2$ & $\select{89.4}\pm0.3$ & $89.3\pm0.2$ & $88.7\pm0.1$ \\
\midrule
\midrule
Dataset &   \multicolumn{4}{c}{\textsc{CIFAR-100 60\% Symmetric}}                                                                                      & \multicolumn{4}{c}{\textsc{CIFAR-100 20\% Asymmetric}} \\
Wait     &                        10 &                       13 &                       16 &                       19               &                         10 &                       13 &                       16 &                       19 \\
\midrule
\name     &                 $55.8\pm0.3$ & $56.6\pm0.5$ & $\select{56.8}\pm0.3$ & $56.7\pm0.3$            &                  $67.6\pm0.1$ & $67.7\pm0.2$ & $\select{67.8}\pm0.2$ & $67.6\pm0.2$ \\
\bottomrule
\end{tabular}

\end{table}

\section{Conclusion and Implications}

%

We propose \name+, a method designed for upweighting or downweighting instances for robust learning in the presence of noise in the labels. \name+ is easy to use as a plug-in substitute for any loss function, \CE{} included. We verify the efficacy of \name+ as well as its natural multi-class extension against noisy labels on the benchmark datasets. This work opens up to a range of new questions. Although margins have been well understood for simpler classification tasks, in the context of deep neural networks  we no longer understand the power of margins. Our work indicates that the margin history (rather than a single margin) is influential in bringing or breaking the stability of deep learning. In addition, this might be exacerbated in the presence of asymmetric noisy labels---a setting particularly powerful for malicious data corruptions.

\section*{Acknowledgments}

This work was supported in part by NSF awards CNS-1730158, ACI-1540112, ACI-1541349, OAC-1826967, the University of California Office of the President, and the University of California San Diego's California Institute for Telecommunications and Information Technology/Qualcomm Institute. We are grateful to CENIC for the 100Gpbs networks.

\bibliography{reference}
\bibliographystyle{icml2021}

\pagebreak
\appendix

\section{Experimental Setup}

\subsection{Binary Case}\label{subsec-exper-binary}

For each dataset and each noise level, we repeat the experiment 10 times using different random seeds. For each seed, we inject random label noise on the training set according to the prescribed noise level, and leave the test set intact.

The hyperparameters of all methods are tuned via 5-fold cross validation. Note that we do not make use of any clean validation sets. In particular, for GCE, $q$ is tuned from $\{0,1,0.3,0.5,0.7,0.9\}$; for SCE, $A$ and $\alpha$ are tuned from $\{-2,-5,-8\}$ and $\{0.01,0.1,1\}$, respectively, using a grid search; for \name, the wait period is tuned from $\{3,4,5,6,7\}$.

For (boot, sneaker) dataset, we use a 4-layer CNN, and train all methods in 70 epochs, where the learning rate decays by a factor of 10 at epoch 50 and 60. For the rest, we use ResNet-18, and train all methods in 100 epochs, unless noted otherwise, where the learning rate decay takes place at epoch 75 and 90. All networks are trained using SGD with momentum 0.9 and weight decay $2\times10^{-4}$. Standard data augmentations (random crop of size 32, random horizontal flip) are applied to CIFAR-10 and CIFAR-100 datasets.

For NPCL, the threshold parameter $C$ is set to $(1-\epsilon)*n$, where $\epsilon$ is the oracle noise level, and $n$ is the batch size. Note that NPCL requires the knowledge of the oracle noise level, whereas the rest of the methods do not have access to this information. In addition, for datasets other than (boot, sneaker), NPCL trains for 200 epochs.

For \name, the initial learning rate is 0.01. The warm-up period is set as in Table~\ref{table-binary-warmup}, and is not further tuned. The selected wait periods are displayed in Table~\ref{table-binary-sym-wait} and \ref{table-binary-asym-wait}.

\begin{table}[ht]
\caption{Warm-up periods of \name{} for different datasets.}
\label{table-binary-warmup}
\tableformat
\begin{tabular}{ccccc}
\toprule
Datasets & \textsc{Boot} & \textsc{Truck} & \textsc{Cat} & \textsc{Aquatic}\\
\midrule
Warm-up Period & 10 & 15 & 20 & 15\\
\bottomrule
\end{tabular}

\end{table}

\begin{table}[ht]
\caption{Wait periods selected on different binary datasets with symmetric label noise.}
\label{table-binary-sym-wait}
\tableformat
\begin{tabular}{ccccccccc}
\toprule
Dataset &   \multicolumn{4}{c}{\textsc{Boot}}                                                                                    &   \multicolumn{4}{c}{\textsc{Truck}} \\
Noise     &                        10\% &                       20\% &                       30\% &                       40\%      &                         10\% &                       20\% &                       30\% &                       40\% \\
\midrule
\name     &                6 & 6 & 5 & 6        &                5 & 3 & 4 & 4 \\
\name+                    &                 6 & 6 & 5 & 6       &                  5 & 3 & 4 & 4\\
\midrule
\midrule
Dataset &   \multicolumn{4}{c}{\textsc{Cat}}                                                                                    &   \multicolumn{4}{c}{\textsc{Aquatic}} \\
Noise     &                        10\% &                       20\% &                       30\% &                       40\%               &                         10\% &                       20\% &                       30\% &                       40\% \\
\midrule
\name     &                 4 & 4 & 5 & 5            &                   4 & 4 & 4 & 5 \\
\name+                    &                 4 & 5 & 5 & 5             &                  7 & 4 & 4 & 6 \\
\bottomrule
\end{tabular}

\end{table}

\begin{table}[ht]
\caption{Wait periods selected on different binary datasets with asymmetric label noise.}
\label{table-binary-asym-wait}
\tableformat
\begin{tabular}{ccccccccc}
\toprule
Dataset &   \multicolumn{4}{c}{\textsc{Boot}}                                                                                    &   \multicolumn{4}{c}{\textsc{Truck}} \\
Noise     &                        (20\%,0\%) &                       (30\%,10\%) &                       (40\%,0\%) &                       (40\%,10\%)      &                         (20\%,0\%) &                       (30\%,10\%) &                       (40\%,0\%) &                       (40\%,10\%) \\
\midrule
\name     &                 6 & 6 & 5 & 6       &                3 & 4 & 3 & 4 \\
\name+                    &                 6 & 7 & 5 & 4       &                  4 & 4 & 3 & 3\\
\midrule
\midrule
Dataset &   \multicolumn{4}{c}{\textsc{Cat}}                                                                                    &   \multicolumn{4}{c}{\textsc{Aquatic}} \\
Noise     &                        (20\%,0\%) &                       (30\%,10\%) &                       (40\%,0\%) &                       (40\%,10\%)                 &                       (20\%,0\%) &                       (30\%,10\%) &                       (40\%,0\%) &                       (40\%,10\%) \\
\midrule
\name     &                 5 & 4 & 4 & 5            &                   4 & 5 & 5 & 6 \\
\name+                    &                  5 & 4 & 4 & 5             &                  4 & 5 & 5 & 6 \\
\bottomrule
\end{tabular}

\end{table}

\subsection{Multi-Class Case (ResNet-32)}\label{subsec-multi-32}

We follow the setup from \citet{pleiss2020}, and consider symmetric and asymmetric label noise on CIFAR-10 and CIFAR-100. In particular, we use ResNet-32 included in the official code of \citet{pleiss2020}, and do not make use of any clean validation sets. For each dataset and each noise level, we repeat the experiment 5 times using different random seeds.

For CIFAR-10, the wait period of \name{} is tuned from $\{4,6,8,10,12\}$, except for symmetric 80\% noise, where it is tuned from $\{10,14,18,22,26\}$. For CIFAR-100, the wait period of \name{} is tuned from $\{4,7,10,13,16,19\}$, except for symmetric 80\% noise, where it is tuned from $\{13,16,19,22,25\}$. The wait period of \name+ is tuned from a neighborhood centered at the selected wait period of \name. For example, if the wait period of \name{} is set to 8, then the wait period of \name+ is tuned from $\{6,7,8,9,10\}$. An exception is CIFAR-10 with symmetric 80\% noise, where the wait period of \name+ is tuned from the same set as that of \name. The hyperparameters of SCE are configured according to the original paper.

For CIFAR-10, we train all methods in 120 epochs with a batch size of 128 and an initial learning rate of 0.1, where the learning rate decays by a factor of 10 at epoch 40 and 80, unless noted otherwise. For CIFAR-100, all methods are run in 150 epochs with a batch size of 128 and an initial learning rate of 0.1, and the learning rate decay takes place at epoch 80 and 120, unless noted otherwise. All networks are trained using SGD with momentum 0.9 and weight decay $2\times10^{-4}$. Standard data augmentations (random crop of size 32, random horizontal flip) are applied.

For \name, the learning rate decay takes place at epoch 80 and 100 or at epoch 100 and 130 for CIFAR-10 or CIFAR-100, respectively. The warm-up period is set to 30 or 40 epochs for CIFAR-10 or CIFAR-100, respectively, and is not further tuned. The selected wait periods are displayed in Table~\ref{table-cifar-sym-wait} and \ref{table-cifar-asym-wait}.

\begin{table}[ht]
\caption{Wait periods selected on CIFAR-10 and CIFAR-100 with symmetric label noise.}
\label{table-cifar-sym-wait}
\tableformat
\begin{tabular}{ccccccccc}
\toprule
Dataset &   \multicolumn{4}{c}{\textsc{CIFAR-10}} &   \multicolumn{4}{c}{\textsc{CIFAR-100}} \\
Noise &                        20\% &                       40\% &                       60\% &                       80\% &                         20\% &                       40\% &                       60\% &                       80\% \\
\midrule
\name                                &                 10 &             8 &             8 &             22 &                  19 &             19 &             16 &             19 \\
\name+                               &                 9 &             7 &             7 &             26 &                  19 &             18 &             17 &             18 \\
\bottomrule
\end{tabular}

\end{table}

\begin{table}[ht]
\caption{Wait periods selected on CIFAR-10 and CIFAR-100 with asymmetric label noise.}
\label{table-cifar-asym-wait}
\tableformat
\begin{tabular}{ccccc}
\toprule
Dataset &  \multicolumn{2}{c}{\textsc{CIFAR-10}} &      \multicolumn{2}{c}{\textsc{CIFAR-100}} \\
Noise &                       20\% &                       40\% &                       20\% &                       40\% \\
\midrule
\name                                &                 8 &             6 &                              16 &             7             \\
\name+                               &                 10 &             6 &                              15 &             6             \\
\bottomrule
\end{tabular}

\end{table}

\subsection{Multi-Class Case (ResNet-34)}\label{subsec-multi-34}

We consider symmetric and class-dependent label noise on CIFAR-10, following \citet{patrini2017}. For each noise level, we repeat the experiment 5 times using different random seeds. For each seed, unlike in previous experiments, we reserve 10\% of the training set for validation, inject random label noise on the remaining training instances, and leave the test set intact. The validation set is considered a held-out set; it is not used as part of the training set.

For \name, the wait period is tuned from $\{3,4,5,6,7\}$, except for symmetric 80\% noise, where it is tuned from $\{5,10,15,20,25\}$. The optimal wait period is selected according to the performance on the validation set. The hyperparameters of SCE are configured according to the original paper.

We use ResNet-34, and train all methods in 120 epochs with an initial learning rate of 0.1, where the learning rate decays by a factor of 10 at epoch 40 and 80, unless noted otherwise. All networks are trained using SGD with momentum 0.9 and weight decay $2\times10^{-4}$. Standard data augmentations (random crop of size 32, random horizontal flip) are applied.

For \name, the learning rate decay takes place at epoch 80 and 100. The warm-up period is set to 40 epochs, and is not further tuned. The selected wait periods are displayed in Table~\ref{table-resnet34-wait}.

\begin{table}[ht]
\caption{Wait periods selected on CIFAR-10 with symmetric and class-dependent label noise.}
\label{table-resnet34-wait}
\tableformat
\begin{tabular}{ccccccccc}
\toprule
Type &   \multicolumn{4}{c}{\textsc{Symmetric}}                                                                                      & \multicolumn{4}{c}{\textsc{Class-Dependent}} \\
Noise     &                        20\% &                       40\% &                       60\% &                       80\%      &                         10\% &                       20\% &                       30\% &                       40\% \\
\midrule
\name     &                 5 & 4 & 5 & 20       &                6 & 6 & 5 & 4 \\
\name+                    &                 6 & 5 & 6 & 25       &                 7 & 6 & 5 & 4 \\
\bottomrule
\end{tabular}

\end{table}

\subsection{Real-World Datasets}\label{subsec-real-world}

We follow the setup from \citet{pleiss2020}, and consider WebVision50 and Clothing100K datasets. In particular, we use ResNet-50 as well as the two datasets included in the official code of \citet{pleiss2020}, and do not make use of any clean validation sets.

We train all methods in 160 epochs with a batch size of 256 and an initial learning rate of 0.1. For \CE, the learning rate decays by a factor of 10 at epoch 80 and 120. For \name, the learning rate decays by a factor of 10 at epoch 120 and 140. All networks are trained using SGD with momentum 0.9 and weight decay $2\times10^{-4}$. Standard data augmentations (random crop of size 224, random horizontal flip) are applied.

On WebVision50, the warm-up period of \name{} is set to 50 epochs. We split 25\% training data into a validation set, and tune the wait period from $\{18,30,42,54,66\}$, and then fine-tune from $\{54,58,62,66\}$. Finally, we set the wait period to 62.

On Clothing100K, the warm-up period of \name{} is set to 70 epochs. We split 25\% training data into a validation set, and tune the wait period from $\{12,24,36,48\}$, and then fine-tune from $\{36,40,44,48\}$. Finally, we set the wait period to 44.
\end{document}